\documentclass{article}
\usepackage{fullpage}
\usepackage{amsmath, amssymb, amsthm, tikz, color}
\usepackage{hyperref}

\newcommand{\beginsupplement}{%
        \setcounter{table}{0}
        \renewcommand{\thetable}{A\arabic{table}}%
        \setcounter{figure}{0}
        \renewcommand{\thefigure}{A\arabic{figure}}%
     }
     
\usepackage[boxed]{algorithm2e}

\DeclareMathOperator*{\argmin}{arg\,min}

\newtheorem{definition}{Definition}

\usepackage{xcolor} % I think this is allowed? If not, we'll remove anyway for final

\title{Fuzzy Simplicial Networks: A Topology-Inspired Model to Improve Task Generalization in Few-shot Learning}
\author{
        Henry Kvinge\thanks{Pacific Northwest National
Laboratory, Seattle, Washington,
USA} ,
        Zachary New\footnotemark[1] ,
        Nico Courts\footnotemark[1] \thanks{Department of Mathematics, University of Washington, Seattle, Washington, USA} ,\\
        Jung H. Lee\footnotemark[1] ,
        Lauren A. Phillips\thanks{Pacific Northwest National Laboratory, Richland, Washington, USA} ,
        Courtney D. Corley\footnotemark[3] ,\\
        Aaron Tuor\footnotemark[1] ,
        Andrew Avila\footnotemark[3] ,
        Nathan O. Hodas\footnotemark[3]\\
}

\begin{document}

\maketitle

\begin{abstract}
Deep learning has shown great success in settings with massive amounts of data but has struggled when data is limited. Few-shot learning algorithms, which seek to address this limitation, are designed to generalize well to new tasks with limited data. Typically, models are evaluated on unseen classes and datasets that are defined by the same fundamental task as they are trained for (e.g. category membership). One can also ask how well a model can generalize to fundamentally different tasks within a fixed dataset (for example: moving from category membership to tasks that involve detecting object orientation or quantity). To formalize this kind of shift we define a notion of “independence of tasks” and identify three new sets of labels for established computer vision datasets that test a model's ability to generalize to tasks which draw on orthogonal attributes in the data. We use these datasets to investigate the failure modes of metric-based few-shot models. Based on our findings, we introduce a new few-shot model called Fuzzy Simplicial Networks (FSN) which leverages a construction from topology to more flexibly represent each class from limited data. In particular, FSN models can not only form multiple representations for a given class but can also begin to capture the low-dimensional structure which characterizes class manifolds in the encoded space of deep networks. We show that FSN outperforms state-of-the-art models on the challenging tasks we introduce in this paper while remaining competitive on standard few-shot benchmarks.

\end{abstract}

\section{Introduction}

Traditionally deep learning requires large amounts of labelled data to build models that do not overfit to their training set \cite{LeCun2015}. However, preparing sufficient amounts of data can be costly, and in many applications impractical, limiting deep learning's utility. To address this challenge, the area of few-shot learning aims to develop methods that leverage the strengths of deep learning to solve problems where one may only have a handful of examples from each class (see \cite{Wang2018} for a survey). 

Many of the most effective models in few-shot learning fall into the family of metric-based methods. Notable examples of such models include Prototypical Networks \cite{snell2017prototypical} and Matching Networks \cite{NIPS2016_6385}. These models rely on an encoder function (usually a deep network) which learns to extract rich features from data while being trained on a related task. At inference time the encoder function maps instances of new classes into the learned feature space and builds a class representation from them. Predictions are then made by comparing the image of unlabeled instances in the encoded space with each of the class representations. Prototypes can be hardcoded as simple geometric structures such as a centroid \cite{snell2017prototypical} or subspace \cite{simon2020adaptive,devos2019subspace} formed by encoded examples, or could be learned via a distance function as in the case of Relation Nets \cite{sung2018learning}.  

While these models have proven to be remarkably successful in many contexts, investigations into their effectiveness have mostly focused on cases where the tasks that the models are evaluated on are broadly similar to those that they trained on (for instance, evaluating the performance of a model on CIFAR100 \cite{krizhevsky2009learning} or Caltech-UCSD Birds 200 \cite{WelinderEtal2010} when it was trained on ImageNet \cite{deng2009imagenet}). In this paper we are interested in understanding how metric-based models handle more challenging tasks with the ultimate goal of understanding how to make them even more responsive and flexible to new examples given at test time. To this end we introduce three new label sets for well-known computer vision datasets. These labels are easy for a human to understand and predict and they draw on many of the same types of features that are useful in class membership tasks (such as edges, texture, and shape). Importantly though, our labels are ``independent'' of the original labels, a notion that we define in Section \ref{sect-orthogonality-of-tasks}. Unsurprisingly, we find that the metric-based models that we evaluated perform very poorly on these tasks that are independent of the biases formed during training on the ImageNet classification task. 

While these results might suggest that our only hope is to re-train the encoder at inference time, we find that even when faced with these challenging label sets the encoder often extracts features that can discriminate between classes. In fact, our analysis suggests that often the class representations themselves fail to capture the relevant features, and that instead unrelated features overwhelm them. This suggests revisiting the kind of representations that we use in our models. Drawing inspiration from a geometric structure known as a simplicial complex, we propose a new model which we call {\emph{Fuzzy Simplicial Networks (FSN)}}. Simplicial complexes can approximate almost all geometric structures arising in nature while at the same time being built from a simple building block: the simplex. Given that they can approximate spaces much more flexibly than centroids or subspaces for example, they are an ideal candidate for class representations.

We show that FSN significantly out-performs other metric-based models (with different representations but the same base encoder architecture) on the challenging label sets we introduce below, after being trained on ImageNet. Following insights into few-shot model evaluation found in \cite{triantafillou2019meta}, we also show that under the same conditions FSN displays strong generalization performance across a diverse range of other datasets.

In summary, our contributions in this paper include the following.
\begin{itemize}
    \item A description of three new label sets for existing computer vision datasets. These new labels allow a few-shot model to be tested on tasks that are independent of the type it was trained for.
    \item We analyze why these datasets are challenging for one of the most popular metric-based few-shot models, Prototypical Networks. 
    \item We introduce a new metric-based few-shot model called Fuzzy Simplicial Networks which models classes as a novel structure called a fuzzy simplicial complex that we define in this paper.
\end{itemize}

\section{Related Work}

\subsection{Few-shot Learning}

There are a number of different approaches to few-shot learning. Fine-tuning methods \cite{chen2019closer} train a model on a surrogate dataset and then fine-tune on a small number of examples. Data augmentation methods \cite{hariharan2017low} produce additional examples of a class through augmentation and other methods. Gradient-based meta-learning \cite{finn2017model,nichol2018first} is a class of sophisticated methods that optimize specifically for model parameters that are easily updated during fine-tuning for each few-shot episode. Metric-based models learn an encoding of the data into a space where the task can be solved using notions of distance or similarity between labeled and unlabeled examples. In this paper we choose to focus on metric-based methods since we are interested in understanding and improving how classes are represented in this encoded space.

\subsection{Episodic Training and Testing}

In this paper we work within the standard few-shot framework \cite{NIPS2016_6385} where a classification task, or {\emph{episode}}, consists of a {\emph{support set}} of labeled examples
\begin{equation*}
    S = \{(x_1,y_1), \dots, (x_r,y_r)\}
\end{equation*}
where $x_i$ is the datapoint and $y_i$ is the label belonging to classes $C = \{c_1,\dots,c_k\}$ and a {\emph{query set}} $Q$ of unlabeled examples also belonging to classes from $C$. 

We write $S_i$ for the subset of $S$ containing only examples with label $c_i$. In this paper we will always assume that $|S_i| = n$ is fixed for all $1 \leq i \leq k$. The number $n$ is known as the {\emph{shots}} of the episode, while $k$ (the number of different classes) is known as the {\emph{ways}}. This is often written as $n$-{\emph{shot}} $k$-{\emph{way}}.

By {\emph{few-shot training}} and {\emph{few-shot evaluation}} we mean the process of iterative training/testing by episode. During training this means doing loss calculation and backpropagation on an episode by episode basis. During evaluation the reported accuracies and losses are averaged across all episodes in the evaluation run. 

\subsection{Metric-based Models} \label{subsect-models}

Metric-based few-shot learning algorithms can be very roughly decomposed into three fundamental components: (i) an encoder function $f_\theta:X \rightarrow \mathbb{R}^{m}$ that takes data from a space $X$ and maps it into a feature (or encoded) space $\mathbb{R}^{m}$, (ii) a method of representing encoded points from a support set class, $f_\theta(S_i)$, as a single coherent representation $\gamma_i$, and (iii) a distance function  $d:\mathbb{R}^m \times \Gamma \rightarrow \mathbb{R}_{\geq 0}$, where $\Gamma$ is the set of all possible representations of the given type associated with the model. Given a query point $q \in Q$, the model predicts $q$ to belong to class $c_t$ if $t = \argmin_{1 \leq i \leq r}d(f_\theta(q),\gamma_i)$.

Probably the most notable example of this type of model is Prototypical Networks \cite{snell2017prototypical} (or ProtoNets). In this case $\gamma_i$ is the centroid of the encoded points $f_\theta(S_i)$, and $d$ is the standard Euclidean distance in the encoded space. Prototypical Networks have proven to be a simple but robust model that has been important in the development of few-shot learning more broadly. In Deep Subspace Networks \cite{simon2020adaptive} and Regression Networks \cite{devos2019subspace}, $\gamma_i$ is a low-dimensional affine subspace that approximates $f_\theta(S_i)$ and $d$ is the usual distance between a point and an affine subspace induced by the Euclidean metric. 

The models above form a single representation for all elements in a support class. As we will show below, there are times where one would like multiple representations for multiple clusters within $f_\theta(S_i)$. A simple approach to this is to set $\gamma_i$ to be the full support set and $d$ to be some flavor of nearest neighbor distance \cite{wang2019simpleshot}. Another more recent approach generalizes Prototypical Networks to allow for multiple centroids to represent a single class \cite{allen2019infinite}.

Finally, in \cite{zhang2018few} $\gamma_i$ is the $k$-simplex whose vertices are formed by the elements of $f_\theta(S_i)$ (see Section \ref{sect-simplicial-complexes} for a refresher on simplices) and $d$ is a distance function obtained by calculating the quotient of the volume of the $(k+1)$-simplex with vertices $f_\theta(S_i)$ and $q$ over the volume of the simplex whose vertices are the points $f_\theta(S_i)$ without $q$. This approach assumes that classes can be represented by a single, connected, convex structure. However, a single simplex does not provide the flexibility for multiple representations within a support class, a feature that was shown to be important in \cite{allen2019infinite}.

The present work attempts to leverage many of the advantages of the above models while avoiding their limitations. In particular, we will chose to work with fuzzy simplicial complexes which can represent a much wider range of geometries than the structures listed above.

\subsection{Topological Data Analysis}
One domain in which simplicial complexes play a leading role is in topological data analysis (TDA). The idea of persistence homology pioneered in \cite{edelsbrunner2000tds} proceeds by computing the homology of a series of Vietoris-Rips complexes (which can be roughly interpreted as simplicial complexes assigned to a point cloud~\cite{vietoris1927}) in order to understand the topology of a dataset. These concepts were further developed and generalized in works such as \cite{zomorodian2005computing} and \cite{carlsson2005}. In order to ensure our models can be trained efficiently, the flavor of simplicial complex we use in this paper differs substantially from those in the TDA literature. To our knowledge the concept of a fuzzy simplicial complex does not currently play a role in TDA.

\section{Independence of Labels}
\label{sect-orthogonality-of-tasks}

In order to better motivate the datasets that we will introduce in this section, we first describe a notion of independence between sets of labels attached to a dataset. 

\begin{definition}
\label{def-label-independence}
Suppose that $\ell:D \rightarrow C=\{c_1,\dots,c_\ell\}$ and $\tilde{\ell}:D \rightarrow \widetilde{C}=\{\tilde{c}_1,\dots \tilde{c}_t\}$ are two labeling functions on a dataset $D$. We say that $\ell$ and $\tilde{\ell}$ are {\emph{independent labelings}} on $D$ if for any randomly chosen $x \in D$, and $c \in C$ and $\tilde{c} \in \widetilde{C}$, 
\begin{equation*}
    p\big(\ell(x)=c, \tilde{\ell}(x)=\tilde{ c}\big)=p\big(\ell(x)=c\big)p\big(\tilde{\ell}(x)=\tilde{c}\big).
\end{equation*}
We say that two tasks $T$ and $\widetilde{T}$ on $X$ are {\emph{independent}} if their corresponding label sets $\ell$ and $\tilde{\ell}$ are independent.
\end{definition}

We now describe three new sets of labels for existing computer vision datasets where each new label set is designed to be (approximately) independent from the original label set. We measure this independence via the metric of mutual information \cite{shannon1948} between the original and new label set in Table \ref{fig-mutual-information}. A pair of variables is independent if and only if the mutual information between the pair is zero. 

We have been careful to choose new label sets that, while approximately independent from the old labels, still depend on the base features such as edges, textures, colors, etc. that the original labels also depended on. Random labels would also be approximately independent, but would be a less direct way of measuring the type of real-world generalization that we are interested in in this paper. We have included additional details such as dataset size and class balance in the Appendix in Section \ref{subsect:new-label-sets}.

\subsection{Stem/No-stem (SNS) Dataset}
\label{subsubsect-stem-no-stem}

\begin{figure*}[t]
\centering
\includegraphics[width=0.9\columnwidth]{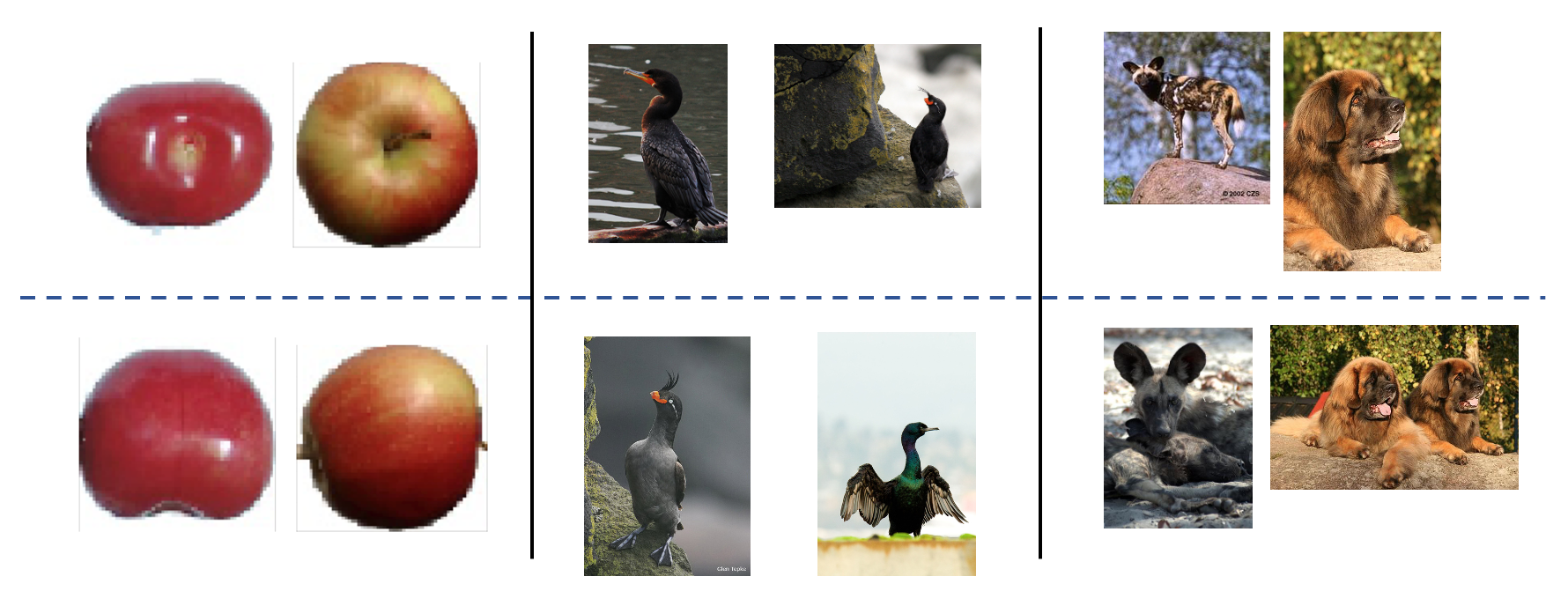} 
\caption{Examples of the stem/no-stem (left), back/no-back (center), and one/many (right).}.
\label{fig-stem-no-stem}
\end{figure*}

The {\emph{Stem/No-stem dataset}} is a re-labeling of the {\emph{Fruits 360 dataset}} \cite{murecsan2018fruit}. The original dataset contains images of different types of fruit with the only variation between images being fruit type and fruit orientation. While the original labels classified images by fruit type, our SNS labels instead focus on orientation and, in particular, whether or not the stem or blossom node are facing the camera (see Figure \ref{fig-stem-no-stem}).

\subsection{Back/No-back (BNB) Dataset}

The {\emph{Caltech-UCSD Birds 200 dataset}} \cite{WelinderEtal2010} is a common benchmark dataset for few-shot learning featuring images of birds. The original labels correspond to the species of bird. The images in this dataset also have a collection of ``parts'' labels corresponding to the position and visibility of different parts of the bird's anatomy (e.g. nape, beak, left wing). We used these attributes to create new labels `Back' and `No-back' for a subset of species based on whether the back of the bird is visible in the image or not.

\subsection{One/Many (OM) Dataset}

The {\emph{Stanford Dogs Dataset}} \cite{KhoslaYaoJayadevaprakashFeiFei_FGVC2011} is another commonly used dataset in computer vision which involves predicting the breed of a dog. The authors of the dataset tag each dog in the image separately, so we were able to extract the number of dogs in each image. We used this information to construct labels `One' and `Many' for a subset of breeds based on the number of dogs in the image.

\begin{table}[t]
\centering
\begin{tabular}{ccc}
Dataset 1 & Dataset 2 & Mutual information\\\hline
Birds & Back/No-back & .043\\
Dogs & One/many dataset & .001\\
Fruits 360 & Stem/No-stem & .031 \\
Fruits 360 & Random binary & .015\\
Fruits 360 & First letter & 2.318 \\
\hline
\end{tabular}
\caption{A calculation of the mutual information between label sets. Random binary consists of random binary labels applied to fruit images in Fruits 360 while `First letter' is a label corresponding to the first letter of the type of fruit pictured in an image. These latter two label sets are introduced as independent and non-independent label sets respectively for reference.}.
\label{fig-mutual-information}
\end{table}

\subsection{What Makes These Datasets Difficult?}
\label{subsect-what-makes-these-datasets-difficult}

In order to understand how to build geometric representations that better capture the underlying structure of a class, we first investigate how a model like ProtoNets can fail on these datasets.

\begin{figure}[t]
\centering
\includegraphics[width=0.49\columnwidth]{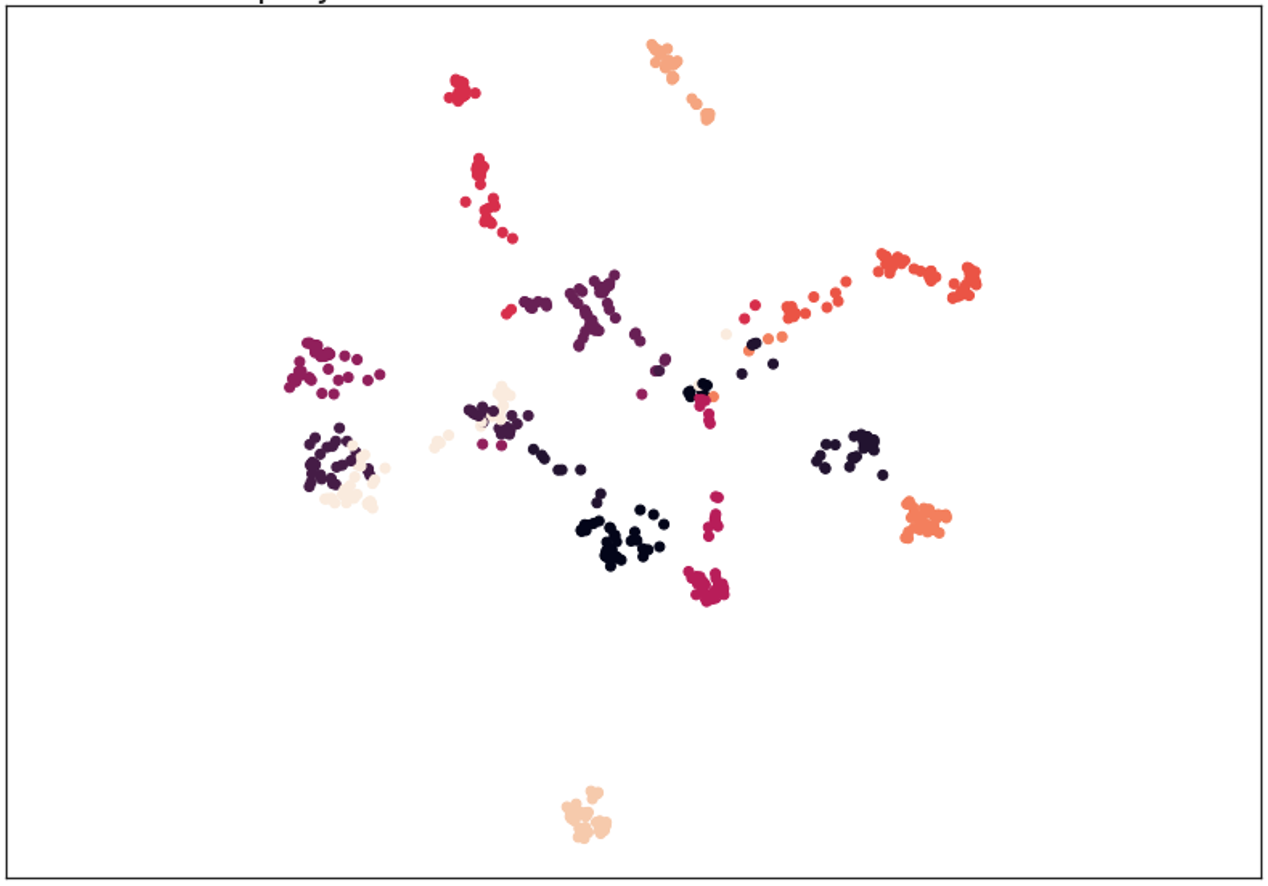}
\includegraphics[width=0.49\columnwidth]{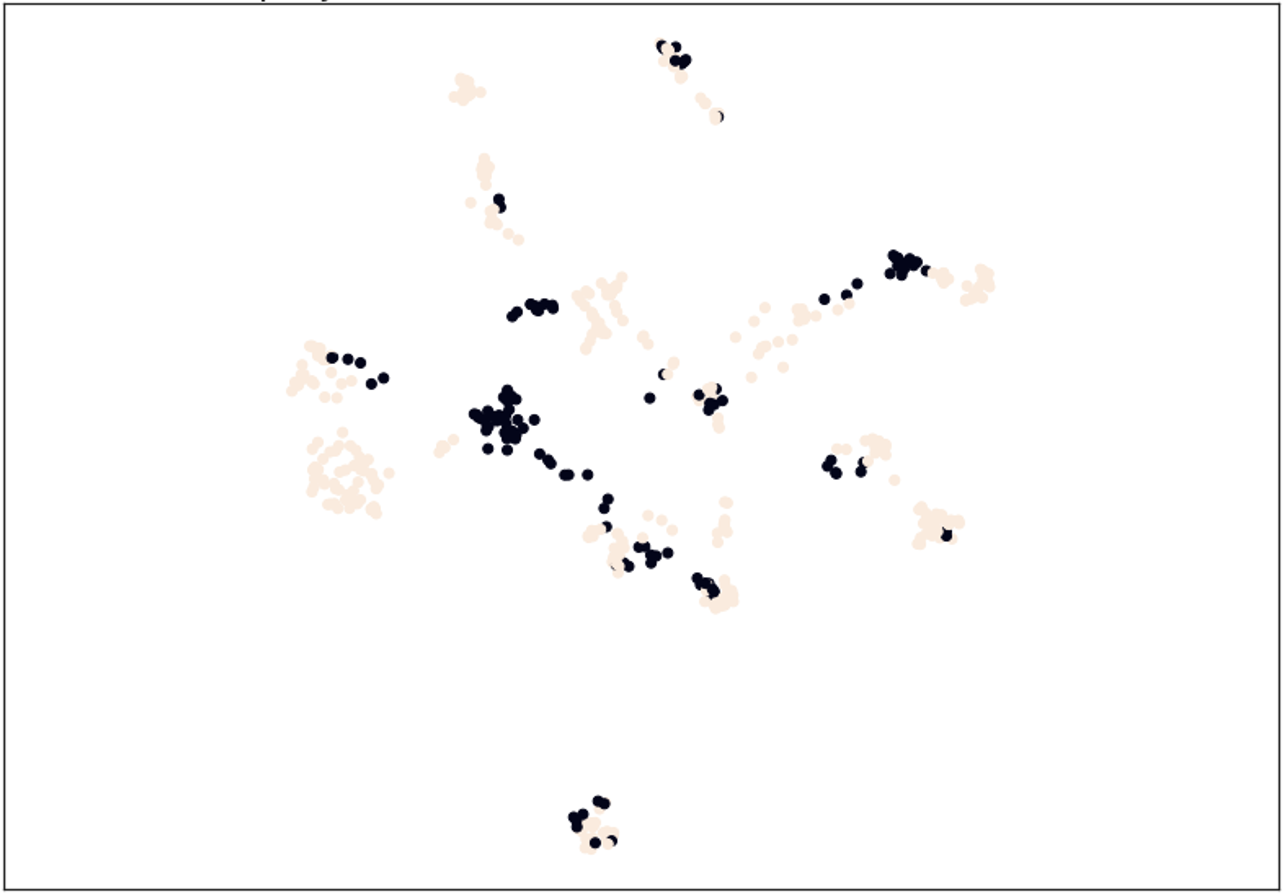} 
\caption{Visualizations of fruit images with the Fruits 360 labels (left) and Stem/No-Stem (SNS) labels (right).
}

\label{fig-feature-space-visualization}
\end{figure}

Figure \ref{fig-feature-space-visualization} contains a visualization of the encoded fruit images that underlie  the Fruits 360  and Stem/No-stem (SNS) datasets. The points are colored by their Fruit 360 labels on the left and SNS labels on the right. The ResNet50 \cite{resnet2016} encoder used to generate these images was episodically trained on ImageNet as part of a ProtoNets model. The visualization suggests that the model has a strong bias toward grouping the images by type of fruit. This can be seen as both a natural consequence of the many visual similarities between images of fruit of the same type and also the ImageNet classification task on which the model was trained.

\begin{table}
    \centering
    {\def\arraystretch{1.5}%
    \begin{tabular}{cc}
         & \textbf{\shortstack{ProtoNet,\\ 5-shot}} \\ \hline
        Fruits 360                          & 96.4\%  \\
        Full SNS                   & 71.2\%  \\
        {SNS, \emph{Apple red 2}}                & 98.7\%  \\
        {SNS, \emph{Green pepper}}                & 99.3\%  \\
        \shortstack{Full SNS with\\ classes mean-centered} & 76.0\% \\\hline
    \end{tabular}}
    \caption{Accuracy of a Prototypical Network model on Fruits 360, SNS, and SNS restricted to particular types of fruit. The {\emph{Apple red 2}} subset and {\emph{Green pepper}} subset contain only images of these fruits with the SNS labels. `Full SNS with classes mean-centered' corresponds to a model where each cluster in the encoded space is mean-centered.}
    \label{fig:SNS_subset_results}
\end{table}

Below we explore an array of additional explanations for the difficulty of these problems and evidence we have collected for or against each:
\begin{enumerate}

\item {\emph{\textbf{Hypothesis:} A ProtoNets model trained on ImageNet does not extract the features required to solve the stem-no-stem problem.}}

%\vspace{.6mm}

The model can easily differentiate between SNS within a particular fruit cluster. In Table \ref{fig:SNS_subset_results} we show the result of evaluating our ProtoNet model on two subsets of the SNS dataset, each of which consists of a fixed type of fruit. As can be seen, the model achieves close to perfect accuracy when restricted to a particular fruit type, (in this case {\emph{Apple red 2}} or {\emph{Green pepper}}). This shows that at least locally (that is, within a cluster), the model extracts high quality features that can be used for discriminating between stem/no-stem images.

\item {\emph{\textbf{Hypothesis:} A ProtoNets model trained on ImageNet is able to extract discriminative features within clusters, but these are not sufficient to differentiate between the classes globally.}} 

%\vspace{.6mm}

The encoding of fruit images obtained from our ProtoNet model (Figure \ref{fig-feature-space-visualization}) is linearly separable with respect to SNS labels. Indeed, using a linear support vector machine model we were able to find a hyperplane in the feature space which separated stem and not-stem points with $100\%$ accuracy. Furthermore, this was not simply a function of the high dimension (2048) of the encoded space. While a random binary labeling of a random sampling of Gaussian points in $\mathbb{R}^{2048}$ with covariance similar to encoded SNS is also linearly separable, the margin is much less significant (a .858 margin for real SNS points and a .004 margin for random labels on random points).

\item {\emph{\textbf{Hypothesis:} The encoder in our ProtoNets model has a strong bias toward extracting features that separate the fruit images by type. This separation tends to overwhelm the features salient to the stem-no-stem task.}} 

%\vspace{.6mm}

To test this we altered our already trained ProtoNets model so that it mean centers all points corresponding to a given type of fruit in the encoded space, removing the separation between clusters. We found that doing this improved the accuracy by nearly $5\%$, indicating that the bias toward separating fruit by type interferes with other tasks.

\item {\emph{\textbf{Hypothesis:} Centroids fail to capture the lower dimensional structure of a class in encoded space.}} 

%\vspace{.6mm}

Using centroids to represent a class makes sense if points from the class actually follow either a Gaussian or some other distribution that has the same intrinsic dimension as the ambient space. The singular values of points from the SNS dataset in the encoded space (see Figure \ref{fig-no-stem-singular} in the Appendix) suggest that the dataset is actually better approximated by a lower dimensional structure. 

In Section \ref{subsect:dimensionality} of the Appendix, we also address whether subspaces can better model the variation in SNS.

\end{enumerate}

The observations above suggest using a more flexible and adaptive framework for building representations which is able to account for multiple representations of a class and also able to model the lower-dimensional structure of the data manifolds on which encoded classes sit.

\section{Simplices, Simplicial Complexes, and Fuzzy Simplicial Complexes}
\label{sect-simplicial-complexes}
Simplicial complexes have a long history in mathematics, and topology in particular, due to the fact that they can effectively approximate a broad range of geometric structures even though they are built from extremely simple constituent parts: simplices. For $k \leq m$ a {\emph{$k$-dimensional simplex}} or {\emph{$k$-simplex}} in $\mathbb{R}^m$, $\Sigma^k$, is the convex hull of $k+1$ (affinely independent) points $x_0, \dots, x_{k} \in \mathbb{R}^m$. 

Simplices of dimensions $0$, $1$, $2$, and $3$ will already be familiar to the reader as points, line segments, triangles, and tetrahedrons. One of the key properties of a $k$-simplex $\Sigma^k$ on vertices $x_0, \dots, x_k$ is that the convex hull of any subset of $\ell+1 \leq k+1$ of these vertices, $x_{i_0}, \dots, x_{i_\ell}$, is itself an $\ell$-simplex known as a {\emph{face of $\Sigma^k$}}. Thus $\Sigma^k$ has, as subsets, $2^{k+1}-1$ nonempty simplices/faces (of dimensions $0$ through $k$) corresponding bijectively to all non-empty subsets of $\{x_0, \dots, x_k\}$. Abusing notation, we write $\Sigma^k = \{x_0, \dots, x_k\}$. The volume of $\Sigma^k$ can be calculated as the square root of the determinant of  $A^TA$ (where $A$ is the matrix whose columns are $x_1-x_0,\dots,x_k-x_0$), normalized by $\frac{1}{k!}$.

Let $\Sigma^k=\{x_0,\dots,x_k\}$ be a $k$-simplex and $q$ any point in its ambient space. We define the {\emph{subspace distance}} $d_\mathrm{sub}(\Sigma^k, q)$ to be the Euclidean distance between $q$ and its projection onto the affine subspace based at $x_0$ and spanned by the vectors $x_1 - x_0,\dots, x_k-x_0$. Note that this definition is invariant under a relabelling of the vertices of $\Sigma^k$.

\begin{figure}
\centering
\includegraphics[width=0.4\columnwidth]{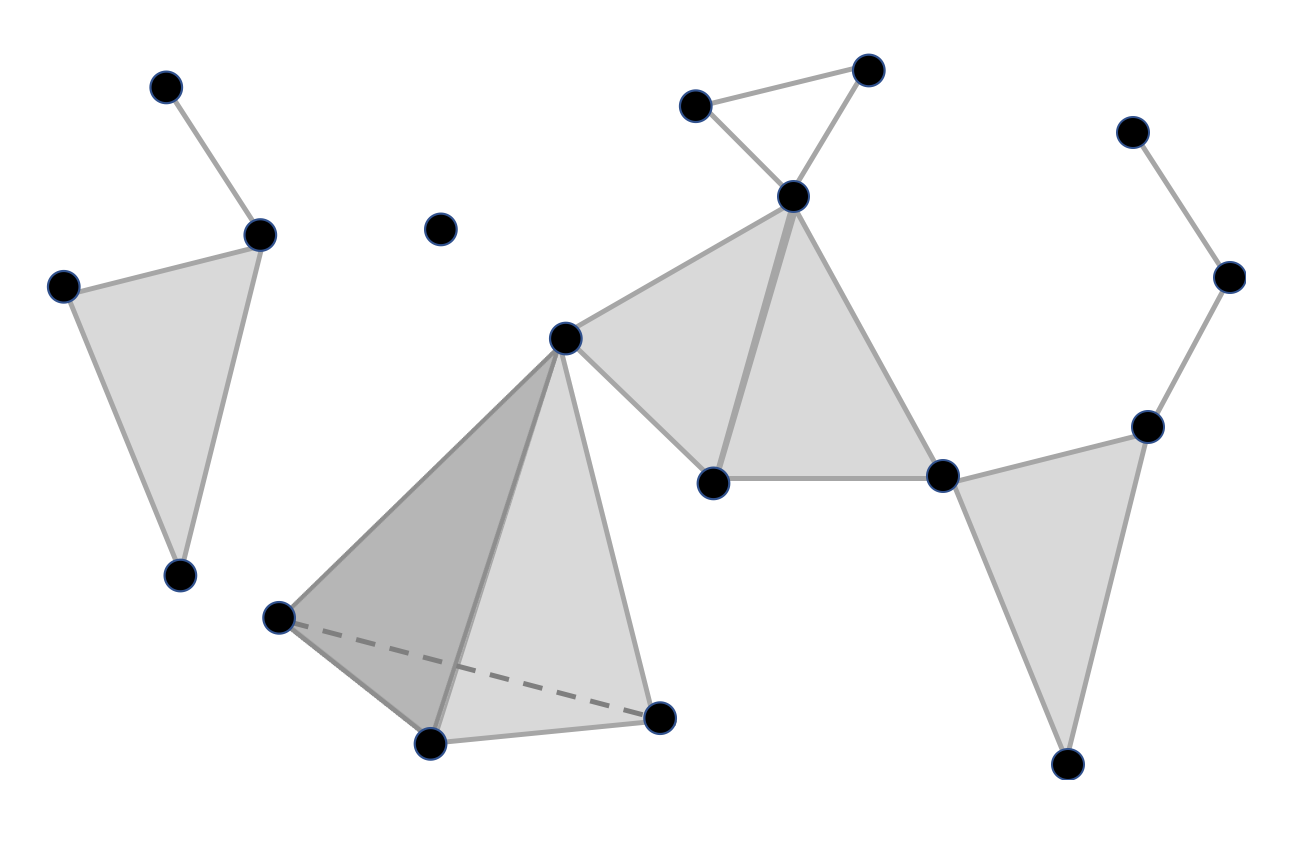}
\caption{An example of a simplicial complex.}
\label{fig:simplices}
\end{figure}

A {\emph{simplicial complex}} $C$ is a collection of simplices of varying dimensions, where individual simplices may be glued together along shared faces (see Figure \ref{fig:simplices} for a visualization). 

A simplex $\Sigma$ in $C$ is called a {\emph{facet}} if it is not a face of a higher dimensional simplex in $C$. For example, the line in the top left corner of Figure \ref{fig:simplices} is a facet even though it is only a $1$-simplex since it is not a face of any higher dimensional simplices.

To adapt simplicial complexes to model real data which can be noisy, we define the notion of a {\emph{fuzzy simplicial complex}} (which was inspired by the use of fuzzy simplicial sets in \cite{mcinnes2018umap}). This in turn was inspired by fuzzy sets, a generalization of sets, where the extent to which an element $x$ belongs to fuzzy set $U$ is measured by a membership function $m:U \rightarrow [0,1]$, with $m(x) = 0$ denoting that $x \notin U$ and $m(x)=1$ corresponding to $x \in U$. 

\begin{definition}
Given a set of points $U = \{x_1, \dots, x_t\}$, a {\emph{fuzzy simplicial complex}} on $U$ denoted by $C =(G(U),m)$ consists of the set $G(U)$ of all simplices that can be constructed from points in $U$ as well as a membership function $m: G(U) \rightarrow [0,1]$ that determines the extent to which each simplex in $G(U)$ belongs to $C$.
\end{definition}

Given that for even a small set $U$, $G(U)$ is very large, in practice we will work with fuzzy simplicial complexes where we assume that facets have fixed dimension $k$, and only work with these simplices in our calculations. We let $G_k(U)$ then denote the set of all $k$-dimensional simplices that can be formed from points in $U$. We then calculate the distance between fuzzy simplicial complex $C = (G_k(U),m)$ and a query point $q$ as:
\begin{equation} \label{eq-fuzzy-simplicial-distance}
    d_{\mathrm{fuzz}}(C,q) := \sum_{\Sigma \in G_k(U)} m(\Sigma)d_{\mathrm{sub}}(\Sigma,q).
\end{equation}
This distance can be interpreted as the sum of the distances from each of the simplices to $q$, weighted by our certainty that the simplex captures the structure of the support set.

\section{Fuzzy Simplicial Networks}
\label{sect-FSN}

In this section we introduce a class of models we call {\emph{Fuzzy Simplicial Networks}} (FSNs). These are metric-based few-shot models which use fuzzy simplicial complexes as representations of support classes. A FSN consists of three components: an encoder function $f_\theta$, a method for building a fuzzy simplicial complex $C_i$ for each encoded support set class $f_\theta(S_i)$, and a method for measuring the distance between an unlabeled query point $f_\theta(q)$ and each $C_i$. We chose the second and third of these so that they are differentiable and the entire model can be trained episodically in an end-to-end manner using backpropogation.

To improve training and inference speed and avoid memory issues, we make free use of the approximations  introduced at the end of Section \ref{sect-simplicial-complexes}. Specifically, a top dimension $k$ is fixed for simplices in all $C_i$ and then all our calculations only include these $k$-dimensional facets in each  $G_k(f_\theta(S_i))$. In particular, we use the $d_{\mathrm{fuzz}}$ distance in \eqref{eq-fuzzy-simplicial-distance} to measure the distance between $C_i$ and query $f_\theta(q)$.

In order to obtain a fuzzy structure on $G_k(f_\theta(S))$ we need a membership function. We define $V_k$ to be the function from all $k$-simplices in $\mathbb{R}^m$ to $\mathbb{R}_{\geq 0}$ such that for a $k$-simplex $\Sigma^k$,
\begin{equation}
V_k(\Sigma^k) := 1/\text{vol}(\Sigma^k).
\end{equation}

We use $V_k$ as the basis for our membership function $m$ under the logic that a simplex with large volume has at least one point that is distant from the others indicating that we should have less certainty that this simplex actually captures the structure of the class. Thus for $\Sigma^k \in G_k(f_\theta(S_i))$ we set
\begin{equation} \label{eqn-membership-function}
    m(\Sigma^k) := \frac{V_k(\Sigma^k)}{\sum_{{\Sigma^k}' \in G_k(f_\theta(S_i))} V_k({\Sigma^k}')}.
\end{equation}
Since we will only be comparing simplices of the same dimension, there are no issues of volumes of different dimension being compared. 

Once the hyperparameter $k$ (the dimension of simplices to be used) has been fixed, the FSN model proceeds with inference as follows. The encoder function $f_\theta$ maps all support set classes $S_1,\dots, S_r$ and query $q$ into the encoded space. For each $S_i$, all $k$-simplices $G_k(f_\theta(S_i))$ are extracted and a membership function $m_i:G_k(f_\theta(S_i)) \rightarrow [0,1]$ as defined above is calculated. The distance function $d_{\mathrm{fuzz}}$ is used to calculate the fuzzy simplicial complex $C_t$ that $f_\theta(q)$ is ``closest to''. Query $q$ is then predicted to belong to class $t$. This algorithm is summarized in Algorithm~\ref{alg:fuzzy}.

\begin{algorithm}
\KwIn{Encoder $f_\theta$, support class sets $S_1, \dots, S_r$, query $q$, simplex dimension $k$.}
\KwOut{The support set index $t$ which $q$ is predicted to belong to.}
\For{$i=1$ to $r$}{
    Compute the set $G_k(f_\theta(S_i))$\;
    Calculate $m_i$ using each $G_k(f_\theta(S_i))$\;
    $C_i \leftarrow (G_k(f_\theta(S_i)),m_i)$\;
    $d_i \leftarrow d_{\mathrm{fuzz}}(C_i,f_\theta(q))$\;
}
$t \leftarrow \argmin_{1 \leq i \leq r}d_i$
\caption{The fuzzy simplicial networks algorithm. $G_k$ is a function that returns all $k$-simplices that can be formed from points from a set, $d_\mathrm{fuzz}$ is the distance from a fuzzy simplicial complex to a point \eqref{eq-fuzzy-simplicial-distance}, and $m$ is the membership function which weights simplices based on the inverse of their volume \eqref{eqn-membership-function}.}
\label{alg:fuzzy}
\end{algorithm}

We note that one could use statistics other than volume to define the membership function $m$. In fact, we also tested models that learned to compute uncertainties from a small fully-connected network that took as input the Gram matrix associated to all simplex vertices shifted to the origin. In general we found that the models where the uncertainty calculation was hard-coded performed better than when uncertainty calculation was learned. We designate this model as {\emph{FSN Learned}} and results of these experiments can be found in the next section.

While our model makes significant gains compared to other few-shot models it also has a few limitations. FSN models take up a larger memory footprint when compared to ProtoNets. FSNs exhibit polynomial memory growth when the number of shots or the number of ways is increased. Additionally, compared to ProtoNets, there is increased complexity in the class representations making it harder to interpret why the model would make a particular choice. FSNs also lose the ability to compare distances between class representations via the same metric used to compute distances between query points.

\begin{table}
    \centering
    {\def\arraystretch{1.5}%
    \setlength{\tabcolsep}{4pt}
    \begin{tabular}{ccccc}
         & \textbf{\shortstack{Stem/\\No-stem}} & \textbf{\shortstack{Back/\\No-Back}} & \textbf{\shortstack{One/\\Many}} \\\hline
        ProtoNet & 73.2$\pm$0.2 & 57.1$\pm$0.3 & 54.8$\pm$0.4 \\
Nearest Neighbor & 74.0$\pm$0.5 & 56.7$\pm$0.2 & 55.5$\pm$0.3 \\
Simplex & 75.4$\pm$0.2 & 57.7$\pm$0.3 & 54.5$\pm$0.2\\
Subspace & 72.7$\pm$0.6 & 57.1$\pm$0.2 & 53.7$\pm$0.2\\
\hline
\textbf{FSN (Ours)} & \textbf{77.9$\pm$0.3} & \textbf{59.2$\pm$0.3} & \textbf{58.0$\pm$0.2}\\
\textbf{FSN Learned (Ours)} & 75.7$\pm$0.7 & \textbf{58.8$\pm$0.2} & 56.6$\pm$0.5\\
\hline
\end{tabular}}
    \caption{Accuracy comparisons across the three challenging label sets introduced above in the $10$-shot, $2$-way regime.}
    \label{fig:model_results_pathological}
\end{table}

\section{Experiments}
\label{sect-experiments}

We are primarily interested in how different class representations can leverage the features extracted from a strong encoder, even when the task they are evaluated on is very different from the one that they were trained for. Thus we trained and validated all the models in our experiments on a few-shot version of ImageNet (see Section \ref{subsect:appendix-experiments} in the Appendix for details). Training was performed in an episodic manner. We trained and tested ProtoNets \cite{snell2017prototypical}, nearest neighbor based models, Simplex \cite{zhang2018few}, Deep Subspace Networks \cite{simon2020adaptive}, and two versions of our FSN (one that uses the hard-coded volume based weighting of simplices and one that learns a weighting as described in Section \ref{sect-FSN}).

We relegate a detailed description of our methods (including the hyper-parameter sweep we did for each model) to Section \ref{subsect:appendix-experiments} of the Appendix, but as an overview, all models used a ResNet50 encoder \cite{resnet2016} with the final layer removed as the base encoder and were initialized (prior to training) with the pre-trained weights available through the TorchVision library \cite{marcel2010torchvision}. Thus the only part of each model that differed was the class representation and distance used in the encoded space. Note that the use of the larger ResNet50 encoder differs from most few-shot learning experiments which leverage smaller encoders \cite{snell2017prototypical,finn2017model,nichol2018first}. We elected to run experiments with a larger encoder to ensure our feature vectors capture as much relevant information from the training task as possible and are not limited by encoder size.

\begin{table*}
    \centering
    {\def\arraystretch{1.5}%
    \setlength{\tabcolsep}{4pt}
    \resizebox{\columnwidth}{!}{\begin{tabular}{cccccccc}
 & \small{Omniglot \cite{Lake1332}} & \small{\shortstack{Adience\\Faces \cite{eidinger2014age}}} & \small{Aircraft \cite{maji13fine-grained}} &  \small{\shortstack{Describable\\Textures \cite{cimpoi14describing}}} & \small{Buildings \cite{phdthesis}} & \small{\shortstack{Fruits\\360 \cite{murecsan2018fruit}}} & \shortstack{Plant\\Seedlings \small{\cite{giselsson2017public}}}\\\hline
ProtoNet & 93.0$\pm$0.2 & 65.7$\pm$0.3 & 55.7$\pm$0.6 & 84.2$\pm$0.2 & 96.2$\pm$0.2 & 99.2$\pm$0.0 & 78.1$\pm$0.7  \\
NearestNeighbor & 89.5$\pm$0.2 & 62.6$\pm$0.3 & 49.1$\pm$0.3 & 76.9$\pm$0.3 & 93.2$\pm$0.3 & 99.4$\pm$0.1 & 74.2$\pm$0.5  \\
Simplex & 91.4$\pm$0.1 & 66.3$\pm$0.3 & 52.7$\pm$0.2 & 82.3$\pm$0.2 & 96.7$\pm$0.2 & 99.6$\pm$0.0 & 80.0$\pm$0.5  \\
Subspace & 91.7$\pm$0.2 & 65.7$\pm$0.4 & 55.0$\pm$0.3 & 83.0$\pm$0.1 & 95.9$\pm$0.2 & 99.6$\pm$0.0 & 78.1$\pm$0.4  \\\hline
\textbf{FSN (Ours)} & \textbf{94.8$\pm$0.3} & \textbf{71.7$\pm$0.3} & \textbf{59.1$\pm$0.2} & \textbf{85.2$\pm$0.2} & \textbf{97.9$\pm$0.1} & \textbf{99.7$\pm$0.0} & \textbf{88.7$\pm$0.5} \\
\textbf{FSN Learned (Ours)} &\textbf{ 94.6$\pm$0.2} & 70.7$\pm$0.2 & \textbf{58.9$\pm$0.4} & \textbf{84.8$\pm$0.2} & \textbf{98.1$\pm$0.1} & \textbf{99.7$\pm$0.0} & \textbf{87.4$\pm$0.4}\\
\hline
\end{tabular}}}
    \caption{Datasets where our models outperform all models evaluated. Accuracies were measured in the $10$-shot, $5$-way regime.}
    \label{fig:best_results}
\end{table*}

\begin{table*}
    \centering
    {\def\arraystretch{1.5}%
    \setlength{\tabcolsep}{4pt}
    \begin{tabular}{ccccccccc}
 & \small{ImageNet \cite{deng2009imagenet}} & \small{CIFAR100 \cite{krizhevsky2009learning}} & \small{\shortstack{CIFAR100\\ Superclass \cite{krizhevsky2009learning}}}& \small{Cars \cite{krause20133d}} & \small{Birds \cite{WahCUB_200_2011}} & \small{Dogs \cite{KhoslaYaoJayadevaprakashFeiFei_FGVC2011}} \\\hline
ProtoNet & 98.2$\pm$0.0 & 84.2$\pm$0.5 & 82.0$\pm$0.3 & \textbf{78.7$\pm$0.2} & \textbf{92.8$\pm$0.1} & 97.2$\pm$0.1   \\
NearestNeighbor & 97.5$\pm$0.0 & 83.7$\pm$0.3 & 80.3$\pm$0.4 & 72.3$\pm$0.3 & 90.2$\pm$0.1 & 96.9$\pm$0.1   \\
Simplex & 97.3$\pm$0.0 & \textbf{86.4$\pm$0.2} & 83.2$\pm$0.2 & 70.5$\pm$0.3 & 88.0$\pm$0.1 & 95.8$\pm$0.1   \\
Subspace & \textbf{98.4$\pm$0.0} & \textbf{86.5$\pm$0.2} & \textbf{84.4$\pm$0.3} & 77.5$\pm$0.4 & 92.5$\pm$0.1 & \textbf{97.5$\pm$0.1}   \\\hline
\textbf{FSN (Ours)} & 98.2$\pm$0.1 & 86.1$\pm$0.1 & \textbf{84.4$\pm$0.3} & \textbf{78.7$\pm$0.4} & \textbf{92.6$\pm$0.2} & 96.9$\pm$0.1   \\
\textbf{FSN Learned (Ours)} & 98.1$\pm$0.1 & 85.9$\pm$0.2 & \textbf{84.0$\pm$0.2} & 77.9$\pm$0.4 & \textbf{92.5$\pm$0.2} & 97.0$\pm$0.0  \\
\hline
\end{tabular}}
    \caption{Datasets where FSN and FSN Learned performed either as well as or less well than other models evaluated. Accuracies were measured in the $10$-shot, $5$-way regime.}
    \label{fig:competitive_results}
\end{table*}

\section{Results}

All models were evaluated a total of 20 times on each dataset. Results reported are means and 95\% confidence intervals computed under the assumption that the data was distributed normally around the true value. The results of tests on our novel label sets are found in Table \ref{fig:model_results_pathological}. Our FSN model using simplex volume to measure membership outperforms all other models we tested, with a maximum margin of 2\% between the confidence intervals for our model and the next best model on One/Many. The fixed FSN also outperforms the variant using a learned membership function, although this model still performs strongly when compared to the non-FSN models.

Although the FSN model performs well on tasks such as SNS, one might wonder whether FSN still performs well on more traditional few-shot learning tasks. To evalutate this, we take our models trained on ImageNet and evaluate them on $13$ datasets. In this way we are able to assess whether the FSN representation also supports generalization to other datasets as in \cite{triantafillou2019meta}. Table \ref{fig:best_results} contains those datasets where FSN did better than all other models. Table \ref{fig:competitive_results} gives the results for those datasets where our models did not outperform others. We note that in all cases our model was within $1\%$ of the accuracy of the of top performing model. While we find that performance on ImageNet itself does not improve, FSN shows a strong advantage on datasets that are very distinct from ImageNet including Adience Faces ($5.4\%$ better), FGVC Aircraft ($3.4\%$ better), and Plant Seedlings ($7.5\%$ better).  More information about each dataset along with their citations is given in Section \ref{sect-testing-datasets} of the Appendix.

Together our results strongly suggest that fuzzy simplicial complexes are a more flexible and adaptive representation that captures the structure of different possible support classes. Our results also suggest that building a membership function based on simplex volume is more robust than trying to learn a weighting from the structure of individual simplices themselves.

\section{Conclusion and Future Work}

In this paper we studied the performance of metric-based few-shot models when they are evaluated on tasks that are significantly different from those that they were trained to solve. Our goal was to find representations that can better model support classes. We showed that even when the encoder function extracts the features needed to distinguish between classes in a support set, the class representatives can fail to capture these. We introduced three new label sets for existing computer vision datasets which are approximately independent from their original labels which we used to evaluate how well models could capture novel class structure. Our analysis of the failures of existing models on these datasets motivated the introduction of our model, FSN. We showed that FSN not only achieves significantly higher average accuracy on the new label sets when compared to a selection of other metric-based few-shot models, but also outperforms or is competitive with these models on common few-shot benchmark datasets.

Our experiments focused on understanding the best geometric structure to use to represent support classes so we did not explore how FSN behaved when it is used in conjunction with other recent approaches such as \cite{bateni2020improved} where the encoder is adapted to the current support at each episode or \cite{Ye2020FewShotLV} where the support set representations are conditioned on one another via a transformer architecture. These would be interesting to explore in future work.

In a different direction, in this work we had to make a number of approximations in order to make simplicial complexes computationally tractable to use in the few-shot setting. While some of these are reasonable from a mathematical perspective (such as the introduction of fuzziness), others feel less justified, such as the restriction to fuzzy-simplicial complexes where facets all have the same dimension. We would like to re-evaluate this approach in the future.

\section{Ethics Statement}

It is increasingly the case that building state-of-the-art machine learning models requires having access to massive amounts of labeled data and computing resources. This sharply limits who has access to the benefits of modern deep learning. We see the present work, and few-shot learning in general, as an effort to broaden access to high-performing deep learning models. 

\bibliographystyle{plain}
\bibliography{egbib}

\newpage

\beginsupplement
\section*{Appendix}
\renewcommand{\thesubsection}{A.\arabic{subsection}}
\subsection{New Label Set Details}
\label{subsect:new-label-sets}

In this section we provide more details about the three new label sets introduced in this paper. 

\subsubsection{Stem/No-stem Dataset}
As mentioned in the main paper, the {\emph{Stem/No-stem dataset}} is a re-labeling of a subset of the {\emph{Fruits 360 dataset}} \cite{murecsan2018fruit}. We used on-site hand labeling to label each image as either `stem' (meaning the stem or blossom node was mostly oriented towards the camera) or `not-stem' (otherwise). Our guiding principle in designing this label set was to identify a property that could be reasonably understood by humans, and was independent of classification of fruit by type.

We avoided fruits where the orientation would be too ambiguous due to their irregular shape, such as bananas; instead we gravitated to mostly-spherical fruits such as tomatoes, apples, grapes, and cherries. The exact fruit classes used, as well as the counts and concentration of `stem' labels can be found in Table \ref{table:stem-no-stem-description}.

\subsubsection{Back/No-back Dataset}

The Caltech-UCSD Birds 200 dataset \cite{WelinderEtal2010} has been used to benchmark few-shot models but typically focuses on species labels. The dataset also includes labels for both `attributes' (e.g. breast color, beak shape) as well as `parts' (e.g. back, beak, nape). It is this latter set of labels that we used to create Back/No-back. 

In this label set, each part of the bird is given a location attribute (as $x,y$-coordinates) as well as a binary `visible' value. We used this final attribute to determine whether the image was `back' or `no-back'. We selected 10 bird species (which are recorded in Table \ref{table:back-no-back-description} along with the relevant statistics) by selecting species that had dramatically different visual appearance--mostly coloring and proportions--and took all available data points on this subset.

\subsubsection{One/Many Dataset}

The Stanford Dogs Dataset \cite{KhoslaYaoJayadevaprakashFeiFei_FGVC2011} has XML files with a separate tag for each dog in a given picture. We extracted the number of tags in each XML file and used this to classify whether the image contained `one' dog or `many'. We restricted ourself to a small subset of all breeds of dogs, which can be seen in Table \ref{table:one-many-description}.

These three breeds were chosen because of their different appearances, as well as the fact that they had similar concentrations of `one' labels. One notices that the mutual information (c.f. Table \ref{fig-mutual-information}) of this labeling with the original dog labels is quite low, likely reflecting the fact that label concentration was actively considered in determining which breeds to use.

\begin{table}[ht]
\centering
\begin{tabular}{ccc}
    Fruit & Count & `Stem' concentration\\\hline
    Tomato 2 & 41 & 0.244 \\
    Papaya & 44 & 0.114 \\
    Apple Red 2 & 45 & 0.289  \\
    Apple Red Delicious & 39 & 0.359 \\
    Pepper Green & 57 & 0.351 \\
    Cherry 1 & 35 & 0.086 \\
    Apple Pink Lady & 47 & 0.362 \\
    Limes & 31 & 0.323 \\
    Cherry 2 & 44 & 0.273 \\
    Pepper Yellow & 33 & 0.273 \\
    Grape Pink & 46 & 0.196 \\
    Pear Red & 38 & 0.237 \\\hline
\end{tabular}
\caption{Stem/No-stem dataset label properties.}
\label{table:stem-no-stem-description}
\end{table}

\begin{table}[ht]
    \centering
    \begin{tabular}{ccc}
        Bird & Count & `Back' concen.\\\hline
Crested Auklet & 44 & 0.568 \\
Pelagic Cormorant & 60 & 0.750 \\
Olive-sided Flycatcher & 60 & 0.367\\
Rose-breasted Grosbeak & 60 & 0.583 \\
Herring Gull & 60 & 0.750 \\
Rufous Hummingbird & 60 & 0.667\\
Tropical Kingbird & 60 & 0.500\\
Arctic Tern & 58 & 0.448 \\
Canada Warbler & 60 & 0.600 \\
Cedar Waxwing & 60 & 0.650 \\\hline
    \end{tabular}
\caption{Back/No-back dataset label properties.}
    \label{table:back-no-back-description}
\end{table}

\begin{table}
    \centering
\begin{tabular}{ccc}
Dog & Count & `One' concentration\\\hline
African hunting dog & 169 & 0.781 \\
Leonberg & 210 & 0.814 \\
Whippet & 187 & 0.781 \\\hline
\end{tabular}
\caption{One/Many dataset label properties.}
    \label{table:one-many-description}
\end{table}

\subsection{The Dimensionality of Classes in Encoded Space}
\label{subsect:dimensionality}
\begin{figure}
\centering
\includegraphics[width=.5\columnwidth]{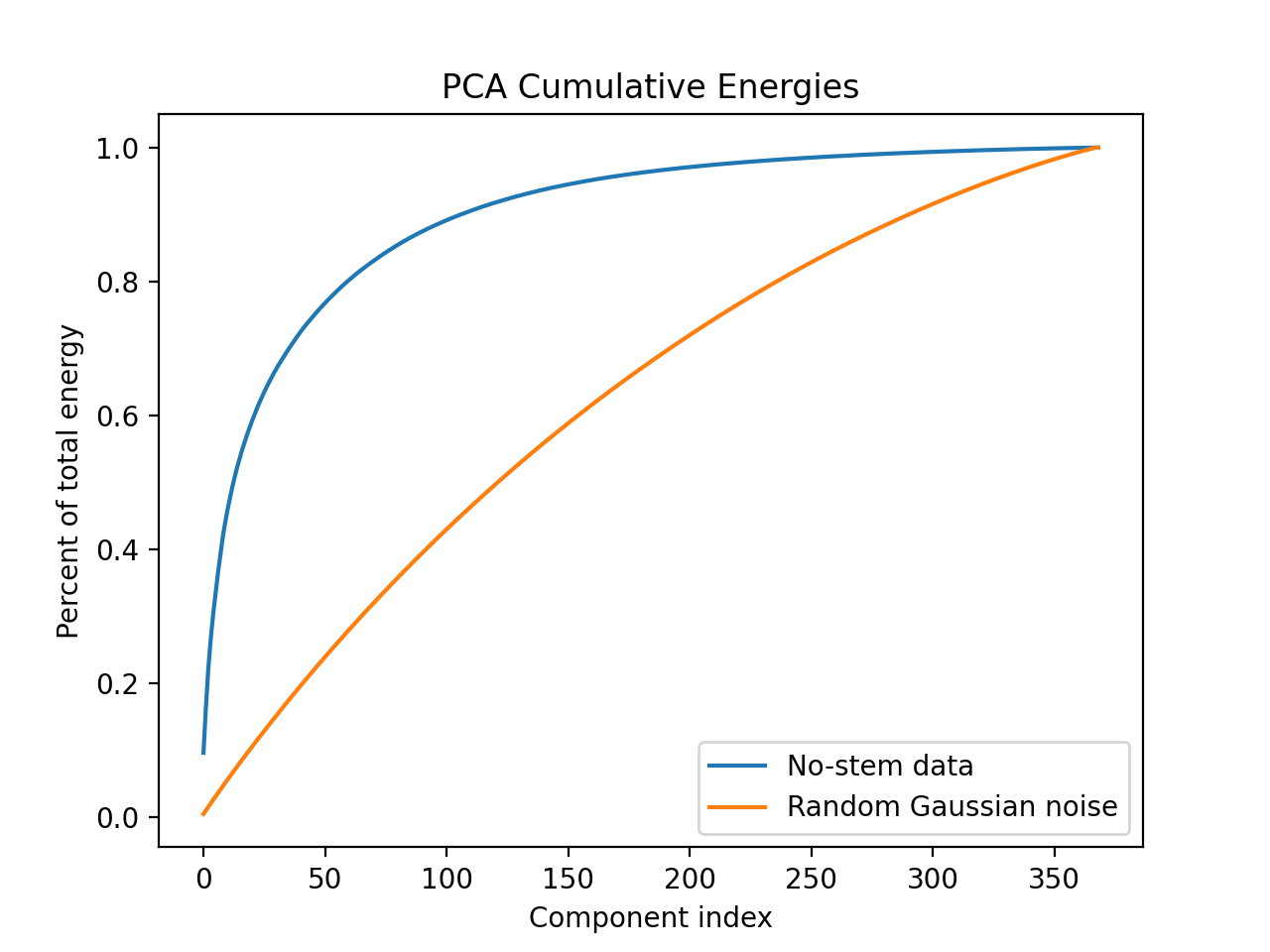}
\caption{The cumulative energy captured by singular values as a function of the number of dimensions for: the collection of encoded and then mean-centered no-stem examples (blue), random Gaussian noise in the same space, $\mathbb{R}^{2048}$ (orange).
}
\label{fig-no-stem-singular}
\end{figure}

One of the motivations for introducing fuzzy simplicial complexes as a representation of support classes was the observation that encoded classes have a lower intrinsic dimension than the ambient encoded space (which in all of our experiments was 2048-dimensional). Figure \ref{fig-no-stem-singular} shows a plot of cumulative energy captured by the singular values as a function of dimension for both the data matrix of all encoded and then mean-centered no-stem examples (blue) and an equal number of random Gaussian points in the same ambient space $\mathbb{R}^{2048}$ (orange). One can see that while the curve for random Gaussian noise is nearly diagonal, suggesting that the distribution that we are sampling from is actually intrinsically $2048$-dimensional, most of the energy for the no-stem class is captured in the first $100$ dimensions, showing that in encoded space the no-stem class is actually a lower dimensional object.

\begin{figure*}
\centering
\includegraphics[width=\columnwidth]{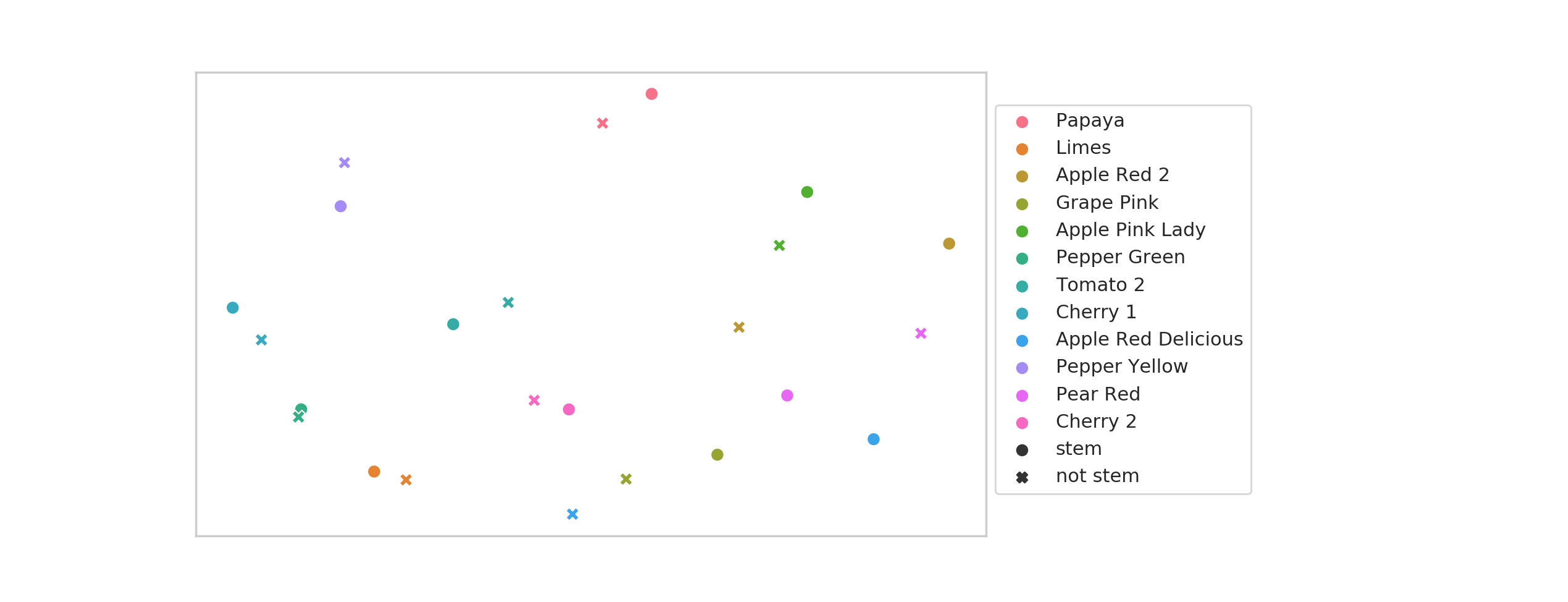}
\caption{A visualization of subspaces (with distances corresponding to the Grassmannian distance \cite{ye2016schubert}) corresponding to different subsets of fruit images organized by fruit type (color) and stem/no-stem (dot or x).
}
\label{fig-grassmann-plot}
\end{figure*}

Alternatively, one could ask if subspaces might be a better way to capture the variation between the stem and no-stem labels of individual types of fruit. To explore this we used a ResNet50 encoder $f_\theta$ trained on the few-shot ImageNet classification task to encode all the images associated with the Fruit 360/SNS datasets. We sorted the encoded points into groups according to type of fruit and whether they belongs to stem or no-stem. As in \cite{simon2020adaptive}, we used PCA to construct a 2-dimensional affine subspace that best captured each of these subsets. Then we calculated the {\emph{Grassmannian distance}} (a metric on subspaces of the same dimension) \cite{ye2016schubert} between these subspaces (translated to the origin) and used multidimensional scaling to visualize them as points in the plane, Figure \ref{fig-grassmann-plot}. 

As can be seen, even as subspaces, sets of points corresponding to the same fruit are much nearer to one another than points corresponding to `stem' or `no stem'. Since nearer points in Grassmannian distance are closer to being parallel, one interpretation of this information is that a two-dimensional subspace representation would struggle capture stem examples from all 12 clusters while simultaneously being distinct from the subspace capturing no-stem. When choosing the dimension of such a subspace representation, one must make a trade-off between larger dimensions, which have a better chance at including a larger part of the support set but may also include noise, and smaller dimensions, which have the opposite properties. So simply scaling up the representation to a larger dimension is not a solution to this problem, and more likely the solution is to use multiple smaller subspace representations. This is, in fact, very similar to what our model does through its use of the subspace (pseudo)metric $d_\mathrm{sub}$.

\subsection{Experimental Details}
\label{subsect:appendix-experiments}

All models were trained with the PyTorch implementation of the Adam optimizer with default values aside from the learning rate. Our first step was to perform a hyper-parameter sweep for each model type (the particular hyper-parameters we evaluated can be found in Table \ref{table-hyperparameter-variables}). The hidden dimension, learning rate, and number of blocks were selected from the best configuration out of 50 runs.

\begin{table}
{\def\arraystretch{1.5}%
\begin{center}
    \begin{tabular}{cc}
        Variable & Quantity\\\hline
        $\ell$ & Learning rate\\
        $d$ & Subspace dimension (PCA components)\\
        $k$ & Simplex dimension\\
        $w$ & Hidden dimension (width of NN)\\
        $b$ & Number of blocks (depth of NN)\\\hline
    \end{tabular}
\end{center}}
\caption{Hyperparameters for our various models and their interpretations. $d$ is relevant to the Subspace models, $k$ is relevant to FSN and FSN Learned. $w$ and $b$ are only relevant to FSN Learned.}
\label{table-hyperparameter-variables}
\end{table}

\begin{table}
{\def\arraystretch{1.5}%
\begin{center}
    \begin{tabular}{cc}
        Variable & Value Ranges\\\hline
        $\ell$ & $10^{-n}$ for $n=1,\dots,6$\\
        $d$ & 1 to 4\\
        $k$ & $1,2,7,8$\\
        $w$ & 256 to 1024\\
        $b$ & 1 to 5\\\hline
    \end{tabular}
\end{center}}
\caption{Hyperparameter values were uniformly randomly sampled from these ranges during our 50-run hyperparameter sweep. Variable interpretations can be found in Table \ref{table-hyperparameter-variables}.}
\label{table-hyperparameter-sweeps}
\end{table}

We selected the final hyperparameters for our tests by comparing how well each model performed on the ImageNet validation set during training. One can find the hyperparameters we selected in Figure \ref{table-chosen-hyperparams}.

\begin{table}
{\def\arraystretch{1.5}%
\begin{center}
    \begin{tabular}{cccccc}
        Model & $\ell$ & $d$ & $k$ & $w$ & $b$ \\\hline
        ProtoNet & $10^{-5}$ & N/A & N/A & N/A & N/A\\
        Simplex & $10^{-5}$ & N/A & N/A & N/A & N/A\\ 
        NearestNeighbor & $10^{-6}$ & N/A & N/A & N/A & N/A\\
        Subspace & $10^{-6}$ & 2 & N/A &N/A  & N/A\\
        FSN & $10^{-6}$ & N/A & 8 & N/A & N/A\\
        FSN Learned & $10^{-6}$ & N/A & 8 & 512 & 1 \\\hline
    \end{tabular}
\end{center}}
\caption{Final hyperparameter selections for our models. Variable interpretations can be found in Table \ref{table-hyperparameter-variables}.}
\label{table-chosen-hyperparams}
\end{table}

After these parameters were determined, each model was then trained four times on our few-shot split of ImageNet using the $10$-shot $5$-way episodic regime for 19,200 episodes. In this split the training and validation sets partition all classes so that instances of each class appear either in training or validation but not both. Of the 1000 classes in the dataset, 800 were randomly selected for training with 100 set aside each for validation and testing. The final weights were retained from each run and used for evaluation. Each evaluation run tested each of the models with the chosen hyper-parameters in the $10$-shot, $5$-way regime if the dataset had $5$ or more classes and $10$-shot, $r$-way regime if the dataset had $r < 5$ classes. In total we evaluated each model against $16$ datasets including the three new label sets introduced in Section \ref{sect-orthogonality-of-tasks} and the test set of our few-shot ImageNet split. 

In our evaluation scheme, the model was tested for 200 episodes and the average accuracy and loss is reported across all these episodes. To provide confidence intervals on our results, we evaluated all four copies of each of our model types five times on each evaluation set, giving a total of 20 experiments per dataset-model pair. These 20 data points were used to estimate the mean and 95\% confidence intervals that appear in the tables in the main paper.

An initial FSN Learned parameter sweep was performed on 12GB NVIDIA Tesla P100 GPUs with access to 16 cores and 64GB of memory. Training runs used for model comparison were all executed on 16GB V100s with access to 64 cores and 500GB of memory. All implementations were created in Python utilizing the PyTorch framework. 

\subsection{The Simplicial Complex Pseudometric}
Although simplifications made in our models (restricting attention to $k$-simplices) obviate the need for comparing distances to simplices of different dimensions, in this section we take the opportunity to discuss how our current framework makes this easy to do. We also discuss how our distance function relates to others in the literature and present a non-fuzzy version of $d_{\text{fuzz}}$. 

The authors of \cite{zhang2018few} provide us with a way to measure distance to a simplex by computing volumes. If $\Sigma$ is a simplex and $q$ an embedded query point, they define the distance between the two to be
\begin{equation}d_\textrm{smplx}(\Sigma,q)=\frac{\text{vol}(\Sigma\cup\{q\})^2}{\text{vol}(\Sigma)^2}.
\end{equation}

We would like a way of efficiently computing the distance between an arbitrary unlabeled point and a simplicial complex. To this end, if $\Sigma^k=\{x_0,x_1,\dots,x_k\}$ is a $k$-simplex and $q$ is a point in the same ambient space as $\Sigma^k$, the {\emph{subspace distance}} $d_\mathrm{sub}(\Sigma^k, q)$ is defined to be the (Euclidean) distance between $q$ and its projection onto the affine subspace based at $x_0$ and spanned by the vectors $x_1 - x_0,\dots, x_k-x_0$. Note that choosing the affine subspace based at $x_i$ for $0 \leq i \leq k$ and spanned by $x_1 - x_i, \dots, x_k-x_i$ gives the same subspace and thus the same distance.

This definition strikes a balance between being easily computable and being a reasonable approximation of the Euclidean distance between a point and a simplex. In fact, when the image falls on the simplex itself (which, in high dimensions, is reasonably likely to happen), the subspace distance is precisely the minimal Euclidean distance between the point and simplex. Furthermore, it can be shown that 
\begin{equation}
    d_\textrm{smplx}(\Sigma^k, q)=\left(\frac{d_\mathrm{sub}(\Sigma^k, q)}{k+1}\right)^2.
\end{equation}
The distances $d_\mathrm{sub}(\Sigma^k, q)$ and $d_\mathrm{sub}(\Sigma^\ell, q)$ are comparable when $k \neq \ell$, whereas $d_\textrm{smplx}$ needs to be scaled in this case.

Because $d_\mathrm{sub}$ is comparable for simplices of different dimensions, we can easily generalize it to a distance between a point and simplicial complex. For simplicial complex $S = \{\Sigma_1, \dots, \Sigma_r\}$ {\emph{simplicial complex subspace distance}} is
\begin{equation}\label{eqn:complex-dist}
    d_\mathrm{cmplx}(C,q) := \min_{\Sigma\subseteq C}d_\mathrm{sub}(\Sigma, q).
\end{equation}
Observe that if $\Sigma'$ is a face of $\Sigma$ then 
\begin{equation}% larger simplices are closer
    d_\mathrm{sub}(\Sigma, q) \leq d_\mathrm{sub}(\Sigma', q) 
\end{equation}
Thus in practice, definition \eqref{eqn:complex-dist} can be simplified to
\begin{equation}
    d_\mathrm{cmplx}(C,q)=\min_{\Sigma\subseteq C_{\mathrm{fac}}}d_\mathrm{sub}(\Sigma, q)
\end{equation}
where $C_{\mathrm{fac}} \subseteq C$ is the set of facets in $C$.

\subsection{Testing Datasets}
\label{sect-testing-datasets}

For these experiments we evaluated the generalization and flexibility of our models by testing on a broad range of publicly available datasets that span a variety of tasks. For datasets where few-shot splits did not already exist we generated new label splits using a 70/15/15 ratio for train/val/test. All evaluation was done on the testing set. We list these datasets below.
\begin{itemize}
    \item Omniglot \cite{Lake1332},
    \item Adience Faces \cite{eidinger2014age},
    \item FGVC Aircraft (Aircraft) \cite{maji13fine-grained},
    \item CIFAR100 \cite{krizhevsky2009learning},
    \item CIFAR100 Superclass \cite{krizhevsky2009learning},
    \item Describable Textures \cite{cimpoi14describing},
    \item Urban Buildings for Image Retrieval (Buildings) \cite{phdthesis},
    \item Fruits 360 \cite{murecsan2018fruit},
    \item Plant Seedlings \cite{giselsson2017public},
    \item ImageNet \cite{deng2009imagenet},
    \item Stanford Cars (Cars) \cite{krause20133d},
    \item Caltech-UCSD Birds-200-2011 (Birds) \cite{WahCUB_200_2011},
    \item Stanford Dogs (Dogs) \cite{KhoslaYaoJayadevaprakashFeiFei_FGVC2011}.
\end{itemize}

\end{document}